# Addressing the interpretability problem for deep learning using many valued quantum logic

Swapnil Nitin Shah (swapnil.nitin.shah@gmail.com)
IBM Systems, 11400 Burnet Road, Austin, Texas

**Deep learning models are widely used for various industrial and scientific applications. Even though these models have achieved considerable success in recent years, there exists a lack of understanding of the rationale behind decisions made by such systems in the machine learning community. This problem of interpretability is further aggravated by the increasing complexity of such models. This paper utilizes concepts from machine learning, quantum computation and quantum field theory to demonstrate how a many valued quantum logic system naturally arises in a specific class of generative deep learning models called Convolutional Deep Belief Networks. It provides a robust theoretical framework for constructing deep learning models equipped with the interpretability of many valued quantum logic systems without compromising their computing efficiency.**

1. Introduction

The last few years have witnessed the rise of machine learning techniques such as Deep Neural Networks (DNNs) for various industrial and scientific applications. The empirical success of Deep Learning (DL) models such as DNNs stems from a combination of efficient learning algorithms and their huge parametric space[1]. Although such models have attained widespread adoption and popularity in the Artificial Intelligence (AI) community, a problem that still remains unsolved is that of explainability or interpretability of such architectures. Deep learning models are inherently opaque, as the exact mechanism by which these generate results for a given set of inputs is unknown. These black box machine learning approaches are in complete contrast with traditional symbolic AI approaches, which offer a straightforward interpretation of their decision process. When decisions derived from such systems ultimately affect human lives (e.g. in medicine, law or defense), there is a growing need for understanding how such decisions are furnished by the underlying AI methods[1].

Post-hoc local explanations and feature relevance techniques are increasingly the most adopted methods for explaining DNNs[1]. Despite having some success, these techniques suffer from two major limitations – (1) Lack of consensus on an objective definition of interpretability for deep learning models and (2) Trade-off between interpretability and performance of these models[1]. A resolution to both of these problems can be achieved if the interpretation of such models can be provided by a formal logic system (as used by symbolic AI approaches) which is directly obtained from the underlying model. This paper demonstrates how a many valued quantum logic system naturally arises in convolutional deep belief networks (CDBNs), which are generative deep learning models that learn a probability distribution over a set of inputs. In particular, the functional renormalization group (FRG) ansatz[2] for CDBNs will be employed for this purpose as described in the following sections.

2. Convolutional Deep Belief Networks

Deep Belief Networks (DBNs) are neural network models comprising of multiple layers of latent (hidden) variables, with each pair of consecutive layers and the connections between them constituting a Restricted Boltzmann Machine (RBM). When trained on samples of data, a DBN learns the probability distribution of the samples. A CDBN comprises of stacked convolutional RBMs with an intermediate non-linear max-pooling layer. A convolutional RBM differs from a regular RBM in that the former introduces a receptive field over the hidden layer, for each spin in the visible layer. A convolution kernel (of the size described by the receptive field) acts on the hidden units to evaluate the probability of the visible units and vice-versa[3].

2.1 The FRG ansatz

The stochastic unsupervised learning problem can be formulated as learning a functional $\Pi$ that encompasses all the stochastic symmetries of the data[2]. This is done by first introducing a bare functional $S[\phi]$ over the set $\Phi$ of data inputs $\phi \in \Phi$ that is invariant under global symmetries of the data. For a probability distribution $P_a[\phi]$, parametrized by the field average $\phi_a$, the constraint that all global symmetries of $S[\phi]$ lead to a corresponding stochastic symmetry of $\Pi[\phi_a]$ (for all fields $\phi_a$), results in a particular form[2] of $P_a[\phi]$

$$P_a[\phi] = \frac{1}{Z} e^{-S[\phi] + \int d^d x \frac{\delta \Pi}{\delta \phi_a}(x) \cdot \phi(x)} \quad (1)$$

where, $Z$ is the partition function given by

$$Z = \int [D\phi] e^{-S[\phi] + \int d^d x \frac{\delta \Pi}{\delta \phi_a}(x) \cdot \phi(x)} \quad (2)$$

$\Pi[\phi_a]$ can then be formulated as the Legendre transform of $\ln(Z)$

$$\Pi[\phi_a] = \int d^d x \frac{\delta \Pi}{\delta \phi_a}(x) \cdot \phi_a(x) - \ln(Z) \quad (3)$$

In general, the effective functional $\Pi$ is not solvable perturbatively in $\phi_a$ owing to divergent integrals. Using tools such as the Functional Renormalization Group (FRG) in Quantum Field Theory (QFT) and making suitable identifications, one can circumvent the problem of divergent integrals in computation of $\Pi[\phi_a]$. The FRG flow can be visualized as an infinite cascade of layers[2] (Fig. 1), where each layer samples a field $\phi$ based on the probability distribution $P_a[\phi]_k$. This distribution is parametrized by (conditioned on) the field $\phi_a$ at the layer above or below (indexed by a continuous index $k$) on moving downwards or upwards in the cascade respectively.

$$P_a[\phi]_k = \frac{1}{Z_k} e^{-S[\phi] + \int d^d x \frac{\delta \Pi_k}{\delta \phi_a}(x) \cdot \phi(x) + T_a[\phi]_k} \quad (4)$$

where

$$T_a[\phi]_k = \int d^d x \, d^d y \left(\phi_a(y) - \tfrac{1}{2}\phi(y)\right) \cdot \hat{R}_k(y-x) \cdot \phi(x) \quad (5)$$

$\hat{R}_k$ is the inverse Fourier transform of the IR regulator $R_k(p)$. It provides a mass-like contribution to the bare action and suppresses the contributions of momenta $p \ll k$. $\Pi_k[\phi_a]$ is the effective average functional at layer with index $k$ that interpolates between effective functional $\Pi$ ($k \to 0$) and bare functional $S$ ($k \to \infty$) according to the FRG flow equation

$$\frac{\partial \Pi_k[\phi_a]}{\partial k} = \frac{1}{2} \int d^d x \, d^d y \left[ \frac{\partial \hat{R}_k(y-x)}{\partial k} \cdot \left( \frac{\delta^2 \Pi_k[\phi_a]}{\delta \phi_a(x) \delta \phi_a(y)}(x,y) + \hat{R}_k(x-y) \right)^{-1} \right] \quad (6)$$

Indeed, using $\phi = \phi_a$ in (4),

$$P_a[\phi_a]_k = e^{-S[\phi_a] + \Pi_k[\phi_a]} \quad (7)$$

From (7), for $k \to 0$,

$$P_a[\phi_a]_0 = e^{-S[\phi_a] + \Pi[\phi_a]} \quad (8)$$

Two consecutive layers in the cascade, differing by an infinitesimal amount $dk$ in their index $k$, form an RBM like structure with the probability of generating a field configuration $\phi$ (on the layer above or below) as given by (4). As shown in Fig. 1, the IR regulator creates a receptive field at each layer, that is inversely proportional to $k$, for computing the change in the effective average functional $\Pi_k[\phi_a]$ as one moves up or down in the cascade. The receptive field transforms regular RBMs into convolutional RBMs (CRBMs) and stacking of CRBMs produces a CDBN. The FRG cascade is henceforth referred to as the FRG ansatz for CDBNs.

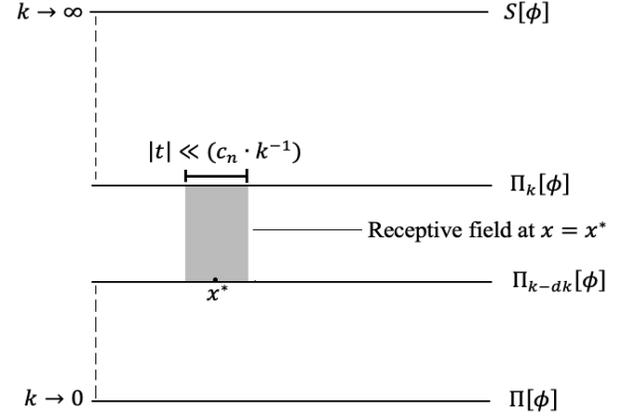

**Figure 1 | Graphical representation of the RG cascade.** The IR cutoff $k$ decreases as one moves down in the cascade recovering the effective functional (action) at the bottom.

In a CDBN, first an input is presented to the layer at the bottom of the stack during the forward pass which is used to generate a probability distribution of the uppermost (hidden) layer conditioned on the input. On the backward pass, a configuration for the hidden spins is sampled from the generated distribution. This is used to generate the probability distribution of the lowermost (visible) layer conditioned on the sampled configuration of hidden spins. The network weights are adjusted so that the probability of generating the original input on the backward pass is maximized. For the FRG ansatz, using equation (8), this is inherently achieved as $\Pi[\phi_a]$ is the Legendre transform of $\ln Z$, which is a convex function[4].

Consider the FRG ansatz with $\phi_a$ as the field configuration of the lowermost layer sampled from a distribution $P_v[\phi]$. Using Bayes' theorem with (7) for consecutive layers $k$ and $k + dk$ in the cascade yields the following

$$P_h[\phi_a]_{k+dk} = P_h[\phi_a]_k \cdot e^{-\frac{\partial \Pi_k[\phi_a]}{\partial k} \cdot dk} \quad (9)$$

Here $P_h[\phi_a]_{k+dk}$ is the probability of generating a field $\phi_a$, marginalized over all field configurations of the layer immediately below (index $k$). Repeatedly applying (9) for the entire cascade,

$$P_h[\phi_a]_\infty = P_v[\phi_a] \cdot e^{\Pi[\phi_a] - S[\phi_a]} \quad (10)$$

For $P_h[\phi_a]_\infty$ to be a probability distribution,

$$\int D\phi_a \, P_v[\phi_a] \cdot e^{\Pi[\phi_a] - S[\phi_a]} = 1 \quad (11)$$

For (11) to be true without restricting the choice of the bare functional $S[\phi]$, one must have (barring the trivial case of $\Pi[\phi_a] = S[\phi_a]$)

$$P_v[\phi_a] = \frac{e^{-\Pi[\phi_a]}}{\int D\phi\, e^{-S[\phi]}} \qquad (12)$$

The above equation (12) completely defines the effective functional $\Pi[\phi]$ in terms of the probability distribution $P_v[\phi]$ over the inputs.

## 3. Quantum Logic in Quantum Computation

Analogous to classical logic gates that operate on binary logic bits (classical memory units), traditionally quantum logic gates operate on qubits (quantum memory units) to perform quantum computation. Even though the energy eigenstates $|0\rangle$ and $|1\rangle$ of the qubits correspond to the classical bit values of 0 and 1, the qubits can also exist in linear superpositions of the two eigenstates. Unlike classical logic gates, which may or may not be reversible, quantum logic gates are always reversible and can be described by unitary operators acting on the tensor product of the Hilbert spaces of the individual qubits[5]. A unitary operator can be represented as a matrix $U$ in any chosen eigenbasis of the product Hilbert space. It satisfies the following:

$$U^\dagger U = UU^\dagger = \mathbb{I} \qquad (13)$$

where $U^\dagger$ is the hermitian conjugate of matrix $U$ and $\mathbb{I}$ is the identity matrix.

As one would expect, quantum logic gates are not just restricted to qubits with 2 eigenstates and offer a straightforward extension to qudits with $d$ eigenstates[5] with gate operations also described by unitary operators. These operators represent $d$-valued quantum logic gates as now they operate on qudits with $d$ possible eigenstates ($d$ possible values).

### 3.1 S-matrix and many valued quantum logic

Fock space is an algebraic construction used in quantum mechanics to construct the quantum state space of an unknown number of identical particles from a single particle Hilbert space. Formally, it is (the Hilbert space completion of) the direct sum of the symmetric or antisymmetric tensors in the tensor powers of a single-particle Hilbert space $H$. A convenient basis for the Fock space is the occupancy number basis. For a choice of an eigenbasis $\{|\psi_i\rangle\}$ ($i$ can be discrete or continuous) of $H$, a state in the Fock space can be represented by the following (or linear combinations thereof)

$$|n_0, n_1, \ldots, n_k\rangle = |\psi_0\rangle^{n_0} |\psi_1\rangle^{n_1} \cdots |\psi_k\rangle^{n_k} \qquad (14)$$

where $n_0, n_1, \ldots, n_k$ are the number of particles in the respective eigenstates. The number of particles in each eigenstate can be 0 and 1 for fermions and 0, 1, 2, … for bosons. As this paper primarily focuses on real scalar fields, the remainder of this discussion is restricted to bosonic systems only.

In QFT, the free (bosonic) particle states are represented by tensor product of single particle momentum eigenstates and can therefore be represented as Fock states (or linear combinations thereof)

$$|n_0, n_1, \ldots, n_k\rangle = |p_0\rangle^{n_0} |p_1\rangle^{n_1} \cdots |p_k\rangle^{n_k} \qquad (15)$$

The S-matrix or scattering matrix is defined as the unitary matrix connecting sets of asymptotically free multi-particle states (the in-states and the out-states) in the Hilbert space of physical states. If the underlying QFT admits a mass gap, the asymptotic states (both in and out states) are Fock states (15)[6].

The S-matrix can therefore be seen as implementing the equivalent of a many valued quantum gate over the in-states to generate the out-states. For bosons, the in and out states can have any (even countably infinite) number of particles for each momentum eigenstate in (15). This provides a natural extension to many valued quantum gates acting on infinite dimensional Hilbert spaces.

## 4. S-matrix from the FRG Ansatz

The S-matrix elements in a QFT with mass gap can be represented in terms of the n-point (time-ordered) correlation functions of the theory using the LSZ reduction formula. These n-point correlation functions can be obtained from the path integral formalism as

$$\langle\Omega|T\{\phi(x_1)\cdots\phi(x_n)\}|\Omega\rangle = \frac{\int D\phi\, \phi(x_1)\cdots\phi(x_n)e^{-S[\phi]}}{\int D\phi\, e^{-S[\phi]}} \qquad (16)$$

The LSZ reduction formula (in momentum space) is as follows[6]

$$\langle p_1, \cdots, p_m | S - \mathbb{I} | q_1, \cdots, q_n \rangle =$$

$$\prod_{i=1}^{m}\left\{\frac{\Gamma(p_i,-p_i)^{-1}}{(2\pi)^{\frac{3}{2}} Z^{\frac{1}{2}}}\right\} \prod_{j=1}^{n}\left\{\frac{\Gamma(q_j,-q_j)^{-1}}{(2\pi)^{\frac{3}{2}} Z^{\frac{1}{2}}}\right\} \cdot \Gamma(\mathbf{p},-\mathbf{q}) \qquad (17)$$

where the left hand side is the interacting S-matrix element (subtracting the identity) and $\Gamma(\mathbf{p},-\mathbf{q})$ on the right hand side is the Fourier transform of the n-point correlation function as given by

$$\Gamma(\mathbf{p},-\mathbf{q}) \equiv \Gamma(p_1,\cdots,p_m;-q_1\cdots,-q_n) =$$

$$\int \prod_{i=1}^{m}\{d^4x_i\, e^{ip_i\cdot x_i}\} \prod_{j=1}^{n}\{d^4x_j\, e^{-iq_j\cdot x_j}\} \langle\Omega|T\{\phi(x_1)\cdots\phi(x_{m+n})\}|\Omega\rangle \qquad (18)$$

$Z$ on the right hand side of (17) is a normalization factor that relates the in-state field to the interacting field. As can be seen from (17), the S-matrix elements are completely defined by the n-point correlation functions (or their Fourier transform).

Next, the n-point correlation functions are computed from the FRG ansatz. Using (10) in (12), the

marginalized probability distribution of field configuration $\phi_a$ in uppermost layer of the cascade is

$$P_h[\phi_a]_\infty = \frac{e^{-S[\phi_a]}}{\int D\phi\, e^{-S[\phi]}} \quad (19)$$

Consider a series of full convolutional layers (infinite receptive field) above the FRG cascade that compute the following (for a choice of $p_1, \cdots, p_n$),

$$G_0(x) = \phi_a(x)$$
$$\vdots$$
$$G_n(x; p_1, \cdots, p_n) =$$
$$\int d^4y\, e^{-ip_n y} G_{n-1}(x; p_1, \cdots, p_{n-1}) G_0(x+y) \quad (20)$$

This is illustrated in Fig. 2 below

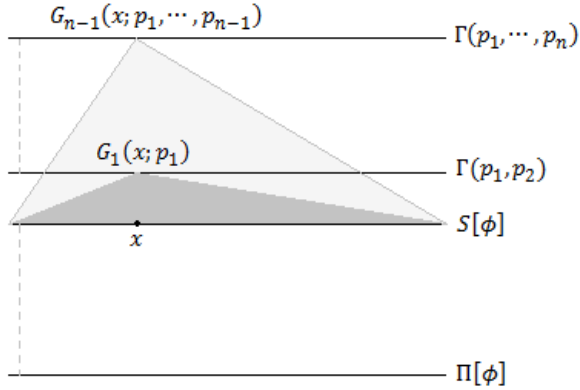

**Figure 2 | Convolutional layers for n-point correlation functions.** Series of full convolutional layers above the FRG cascade compute the n-point correlation functions.

From (16), (18), (19) and (20),

$$\Gamma(p_1, \cdots, p_n) = \int D\phi_a\, P_h[\phi_a]_\infty H_n[\phi_a] \quad (21)$$

where

$$H_n[\phi_a] = \int d^4y\, e^{-ip_n y} G_{n-1}(y; p_1, \cdots, p_{n-1}) \quad (22)$$

If $\phi_a^r$ represents the $r^{\text{th}}$ field generated at the uppermost layer of the cascade (from the probability distribution (19)), it follows from the law of large numbers and (21) that

$$\lim_{r \to \infty} \frac{1}{r} \sum_{i=1}^{r} H_n[\phi_a^i] \approx \Gamma(p_1, \cdots, p_n) \quad (23)$$

The steps outlined above provide a complete prescription of how the n-point correlation functions (and consequently the S-matrix elements) can be computed from the FRG ansatz.

The n-point correlation functions can also be obtained from the partition function (2) as

$$\langle \Omega | T\{\phi(x_1) \cdots \phi(x_n)\} | \Omega \rangle = \frac{1}{Z} \frac{\delta}{\delta J(x_n)} \cdots \frac{\delta}{\delta J(x_1)} Z[J] \Big|_{J=0} \quad (24)$$

where

$$J(x) = \frac{\delta}{\delta \phi(x)} \Pi[\phi] \quad (25)$$

The n-point correlation functions thus uniquely define the partition function (as $Z[J]$ is convex analytic). Also, (12) uniquely defines the effective functional $\Pi[\phi]$ in terms of the input distribution $P_v[\phi]$. As $\Pi[\phi]$ and $\ln Z[J]$ are connected by Legendre transform (3), a bijection is established between the n-point correlation functions (consequently the S-matrix) and the input distribution $P_v[\phi]$ as described below

S-matrix $\leftrightarrow \Gamma(p_1, \cdots, p_n) \leftrightarrow Z[J] \overset{\phi}{\longleftrightarrow} \Pi[\phi] \leftrightarrow P_v[\phi]$

This establishes a one-to-one correspondence between the S-matrix elements and the input distribution $P_v[\phi]$ for an input field $\phi$. In other words, the input distribution $P_v[\phi]$ can be completely determined from the S-matrix elements and vice-versa. This shows not only how a many valued quantum logic system naturally arises in the FRG ansatz, but also completely defines the input probability distribution $P_v[\phi]$ in terms of the S-matrix elements representing the quantum logic system.

## 5. Conclusion

For a long time, researchers have sought to develop AI approaches that have the learning efficiency of neural network models with the interpretability of symbolic AI approaches or formal systems. While both these schools of thought have existed independently since the beginning of AI, there has been very little research on approaches that can benefit from a combination of both. Deep learning models are being employed for all industrial and scientific applications with considerable success. These models are becoming increasingly complex with a multifold increase in their parametric space brought about by the advent of better hardware and software capabilities of modern-day computing systems. Research on using quantum computation for machine learning has also picked up pace, adding to the complexity of these models. With this increasing complexity, however, the interpretability of these models (how a particular decision is made by the algorithm in response to the input) has become an extremely challenging problem to solve. When these complex models are employed for crucial decision making (such as in medicine, defense, etc.), it is essential for the human user to understand the rationale behind the decisions made by the model in order to ensure fidelity of outcomes. While some attempts have been made to provide an explanation for these decisions, they often suffer from lack of an objective definition of interpretability resulting in ambiguity. Also, most of these approaches require a trade-off between how interpretable the resulting model is and its computing efficiency.

This paper utilizes concepts from machine learning, quantum computation and QFT to demonstrate how a many valued quantum logic system (identified with the S-matrix in QFT) naturally arises in the FRG ansatz of CDBNs. To do this, first the input probability distribution $P_v[\phi]$ is described as a function of the effective functional $\Pi[\phi]$ in the FRG ansatz. This is followed by an extension of multi-valued quantum logic to infinite dimensional Hilbert spaces by means of the S-matrix in QFT. Then, it is shown how S-matrix can be computed from the FRG ansatz using a series of convolutional layers above the FRG cascade. Finally, a one-to-one correspondence between the input probability distribution $P_v[\phi]$ and the S-matrix elements is established (mediated by the Legendre transformation from the effective functional $\Pi[\phi]$ to the log-partition function $\ln Z$ and vice-versa). This paper thus provides a robust theoretical framework for constructing deep learning models equipped with the interpretability of many valued quantum logic systems without compromising their computing efficiency.